\documentclass[11pt,a4paper]{article}
\usepackage[hyperref]{naaclhlt2018}
\usepackage{times}
\usepackage{latexsym}
\usepackage{url}
\usepackage{graphicx}
\usepackage{caption}
\usepackage{subcaption}
\usepackage{multirow}
\usepackage{arydshln}

\usepackage{amsmath}

\aclfinalcopy 

\title{PlusEmo2Vec at SemEval-2018 Task 1: Exploiting emotion knowledge from emoji and \#hashtags}

\author{Ji Ho Park, Peng Xu, Pascale Fung \\
  Centre for Artificial Intelligence Research (CAiRE) \\
  Hong Kong University of Science and Technology \\
  {\tt \{jhpark, pxuab\}@connect.ust.hk},  {\tt pascale@ece.ust.hk}} 
\date{}
\begin{document}
\maketitle
\begin{abstract}
  This paper describes our system that has been submitted to SemEval-2018 Task 1: Affect in Tweets (AIT) to solve five subtasks. We focus on modeling both sentence and word level representations of emotion inside texts through large distantly labeled corpora with emojis and hashtags. We transfer the emotional knowledge by exploiting neural network models as feature extractors and use these representations for traditional machine learning models such as support vector regression (SVR) and logistic regression to solve the competition tasks. Our system is placed among the Top3 for all subtasks we participated. 
\end{abstract}

\section{Introduction}

Finding a good representation of texts is very challenging since texts are sequences of words which are represented in a discrete space of the vocabulary. For this reason, many past works have investigated in finding the mapping of words \cite{mikolov2013distributed,pennington2014glove} or sentences \cite{kiros2015skip} to continuous spaces, so that each text can be represented by a fixed-size, real-valued N-dimensional vector. This vector representation then can be applied to machine learning models to solve problems like classification and regression. A good representation should contain essential information inside each text and be a useful input for statistical models.

Emotions in texts further deepen the complexity of modeling natural language since they not only depend on the semantics of a language but also are inherently subjective and ambiguous. Despite the difficulty, accounting for emotion is important in achieving true natural language understanding, especially in areas involving human-computer interactions such as dialogue systems \cite{fung2015robots}. 

Humans can naturally capture and express different emotions in texts, so machines should also be able to infer them. Many works \cite{tang2014learning,felbo2017using,thelwall2017heart} explored modeling sentiment or emotion in texts in various directions. This work is highly related to these efforts.  

Semeval-2018 Task 1: Affect in Tweets (AIT-2018) encourages more efforts in this area with the task of sentiment analysis, which is one of the most practical applications of modeling emotional text representations. We have participated in five subtasks regarding English tweets: emotion intensity regression, emotion intensity ordinal classification, valence (sentiment) regression, valence ordinal classification, and emotion classification (More details on the tasks in \citet{SemEval2018Task1}). 

Although these five tasks take different formats, the most important objective is finding a good representation of the tweets regarding emotions. However, the given competition training datasets are too small to achieve our goal (Table \ref{table:dataset}). Therefore, we explore utilizing larger datasets that are distantly supervised by emojis and hashtags to learn a robust representation and transfer the knowledge of each dataset to the competition datasets to solve the tasks. We aim to minimize the use of lexicons and linguistic features by replacing them with continuous vector representations.

\section{Emoji sentence representations}
 \label{sec:emoji}
Thanks to the endless stream of social media such as Twitter and Facebook, researchers nowadays are lucky enough to have access to almost an unlimited number of texts generated every day. Nevertheless, annotating these texts with explicit emotion or sentiment human labels is very expensive and difficult. For this reason, many works naturally focused on finding direct or indirect evidence of emotion inside each text, such as hashtags and emoticons \cite{suttles2013distant,wang2012harnessing}, and found them useful to distantly label an emotion of each text. Furthermore, the recent popular culture of using emojis \cite{wood2016ruder} inside social media posts and messages provides us even richer evidence of different emotions, and they have been proven to be very effective in learning rich representations for various affect-related tasks \cite{felbo2017using}.

\begin{figure}[t]
\centering
\includegraphics[width=6cm]{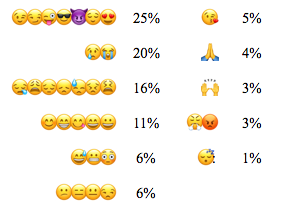}
\caption{11 clusters of emojis used as categorical labels and their distributions in the training set. Because some emojis appear much less frequently than others, we group the 34 emojis into 11 clusters according to the distance on the correlation matrix of the hierarchical clustering from \citet{felbo2017using} and use them as categorical labels}
\label{fig:emoji_dist}
\end{figure}

\subsection{Methodology \& Emoji Dataset}
In this paper, we compare two models using two different emoji dataset to transform the competition data into robust sentence representations.

First model is the pre-trained DeepMoji model \cite{felbo2017using}, which is trained through emoji predictions on a dataset of 1.2 billion tweets with 64 common emoji labels. We use the pretrained deep learning network, which consists of Bidirectional Long Short Term Memory (Bi-LSTM) with attention, except the last softmax layer, as a feature extractor of the original competition datasets. As a result, each sample is transformed into a 2304-dimensional vector from the model.

The second model is our proposed emoji cluster model. We crawled 8.1 million tweets with each of which has 34 different facial and hand emojis, assuming these kinds of emojis are more relevant to emotions. Since some emojis appear much less frequently than others, we cluster the 34 emojis into 11 clusters (Figure \ref{fig:emoji_dist}) according to the distance on the correlation matrix of the hierarchical clustering from \citet{felbo2017using}. Samples with emojis in the same cluster are assigned the same categorical label for prediction. Samples with multiple emojis are duplicated  in the training set, whereas in the dev and test set we only use samples with one emoji to avoid confusion. We then train a one-layer Bi-LSTM classifier with 512 hidden units to predict the emoji cluster of each sample. We take part of the dataset to construct a balanced dev set with 15,000 samples per class (total 165,000) for hyperparameter tuning and early stopping. We use 200 dimension Glove vectors pre-trained on a much larger Twitter corpus to initialize the embedding layer.

The motivation for exploring two different models is that, firstly, we want to replicate the effectiveness of using emoji for representing emotions from the previous work \cite{felbo2017using} with a smaller dataset and a simpler model. Note that the dataset size of the emoji cluster model is less than 1\% of that of the first model, whereas DeepMoji uses more than 1 billion training samples. Moreover, the first model implements a two-layer Bi-LSTM with self-attention, which has much more parameters than the second model's simple one-layer Bi-LSTM does. Secondly, we want to verify that ensembling both emoji representations trained from different datasets to boost our performance. We will present the result of the comparisons and the ensembles in Section \ref{sec:reg-result}.

\begin{table}[h]
\small
\begin{center}
\begin{tabular}{|l|}
\hline \bf One thing i dislike is laggers man \\ 
\hline
I hate inconsistency \\
The paper is irritating me \\
As of right now i hate dre \\
\hline \bf im sick of crying im tired of trying \\ 
\hline
why body pain why \\
uuugh i really have nothing to do right now \\
i dont wanna go back to mex \\
\hline \bf looking forward to holiday \\ 
\hline 
well today am on lake garda enjoying the life \\
perfect time to read book \\
im feeling great enjoying my holiday\\
\hline
\end{tabular}
\end{center}
\caption{\label{table:emoji-eval} Test samples from the Emoji Cluster model and their top-3 nearest sentences according to the learned representations. It shows that emotionally similar sentences are clustered together }
\end{table}

\subsection{Evaluation}

As a result, the model can achieve 29.8\% top-1 accuracy and 61.0\% top-3 accuracy on the emoji cluster prediction task. Since the objective of this model is not to predict the cluster label but to find a good sentence representation, we visualize the test set samples to discover that samples with similar semantics and emotions are grouped together (Table \ref{table:emoji-eval}). Finally, similar to the first model, we use this model as a feature extractor on the competition datasets. Each text sample in the competition datasets is transformed into a 512-dimensional vector through the model except the last class predicting softmax layer.

In conclusion, we trained two deep learning models with two different emoji datasets to extract emoji representations of the competition datasets. They are transformed into high dimensional, real-valued, and continuous vectors, which can be used as features for the classification and regression tasks. 

From now on, we will call the vectors from the first model, \textit{DeepMoji representations}, and those from the second, \textit{Emoji Cluster representations}.

\section{Emotional word vectors (EVEC)}
\label{sec:emo2vec}

We also explore word-level representations, along with emoji sentence representations. Although sentence-level representations already build up from word representations (in particular we use pretrained Glove vectors \cite{pennington2014glove}), they may not be enough to attend to the valence that each word contains. Previous works {\cite{tang2014learning}} examine the significance of using sentiment-specific word embedding for related tasks. For this reason, we train emotional word vectors that not only cluster together direct emotion words such as “anger” and “joy”, but also capture emotions inside indirect emotion words, such as anger inside “headache” and joy inside “beach”. We learn these vectors by training a Convolutional Neural Network (CNN) from another separate Twitter corpus distantly labeled with hashtags.

\subsection{Methodology}
Our intuition to learn effective emotion word vectors is that given a document labeled with emotion there exists one or more emotionally significant words inside. Nevertheless, we do not know which ones are more important. We assume that a deep learning model, which learns the representations of the data with different level of abstractions \cite{Lecun2015} will be able to capture those words and encode the information in its word embedding layer while classifying the document’s label. 

For the model structure, we use CNN since it is proven to be effective in text classification tasks by looking at the document’s n-gram features and its gradient can be directly back-propagated to the word embedding, whereas Recurrent Neural Network (RNN) models are updated sequentially. We use a similar structure used by \cite{Kim2014}, which includes a max-pooling layer to force the network to find the most relevant feature for predicting the emotion class correctly.

After the CNN network learns how to classify the documents into different emotion categories, we extract emotional word vectors from the network's embedding layer and use them as same as how other word embeddings, such as word2vec \cite{mikolov2013distributed} or Glove, are used, treating them as features for other classification or regression models.

\begin{table*}[h]
\small
\begin{center}
\begin{tabular}{|l|l|l|}
\hline Emotion Label & \% & Samples  \\ 
\hline Joy & 36.5\% & It's been such a great week \bf \#happy \\ 
 Sadness & 33.8\% & I think I miss my boyfriend \bf  \#lonely \\ 
 Anger & 23.5\% & Ignoring me isn't going to make our problems go away. \bf  \#annoyed \\
 Fear & 6\% & What to wear for this job orientation.. \bf  \#nervous \\ \hline

\end{tabular}
\end{center}
\caption{\label{table:description-db} Description of the Twitter hashtag corpus. Hashtags at the end were removed from the document and used as labels. It is hard to construct a well-balanced dataset for all four classes since Twitter users tend to use more hashtags related to happy and sad emotions.}
\end{table*}

\subsection{Hashtag Dataset}

To accumulate a large corpus of emotion-labeled texts, we use a distant supervision method by using hashtags of tweets to automatically annotate emotions. Such method has proven to provide relevant emotion labels by previous works \cite{wang2012harnessing}. Their source of the emotion words came from emotion words list made from ~\newcite{Shaver1987}, where the authors organize emotions into a hierarchy in which the first layer contains six basic emotions and each emotion has a list of emotion words. \citet{wang2012harnessing} again expanded the list by including their lexical variants and also introduced some filtering heuristics, such as only using tweets with emotional hashtags at the end of tweets to make the distant supervision more relevant to human annotation. We combine their dataset, another public dataset \footnote{http://hci.epfl.ch/sharing-emotion-lexicons-and-data\#emo-hash-data}, which used the same method, and our own extracted tweets between January and October 2017 using the Twitter Firehose API.

For the emotion labels, we focus on four emotion categories: joy, sadness, anger, and fear, since the competition tasks are only limited to those categories. In total, our hashtag dataset consists of 1.9 million tweets (Table \ref{table:description-db}).

\subsection{Evaluation}

For every sample in the SemEval competition dataset, we extract all emotional word vectors of the words in the sentence and simply average them into one vector. For words out of vocabulary of the hashtag corpus, we add zero vectors with the same dimension. As a result, every sentence is transformed into a 300-dimension vector to be used as features for the competition tasks. We expect these emotional word vectors can replace sentiment or emotion lexicons, since they are continuous representations learned from a large corpus, which can be more robust and rich in information about emotions inside words.

\section{System Description}
\subsection{Features}

These are the three features that are used as input for our system to solve SemEval-2018 Task 1.

{\bf Emoji Sentence Representations}: Two models will be compared - \textit{DeepMoji representations} (2304 dimensions) and \textit{Emoji cluster representations} (512 dimensions). See Section \ref{sec:emoji}.

{\bf Emotional Word Vectors (EVEC)}: Average of emotional word vectors learned from hashtag corpus (300 dimensions). See Section \ref{sec:emo2vec}.

{\bf Tweet-specific features}: We employ Tweet-specific features to capture information that two previous representations cannot. Inspired from the previous SemEval papers \cite{zhou2016ecnu,balikas2016twise}, we choose five features, (1) number of words in uppercase, (2) number of positive and negative emoticons, (3) Sum of emoji valence score \footnote{https://github.com/words/emoji-emotion}, (4) number of elongated words, and (5) number of exclamation \& question marks. Note that we do not use any linguistic features or sentiment/emotion lexicons for our system.

\subsection{Preprocessing}
Tweets in the competition datasets are tokenized after all non-alphanumeric characters are removed, except for extracting tweet-specific features. Some words, especially for hashtags, are merged together (e.g. \#iloveyou), so unknown words in the vocabulary is put into a wordsegment library \footnote{http://www.grantjenks.com/docs/wordsegment/} to preserve the right segment (e.g. “i”, “love”, “you”). Then, the tokens are transformed into emoji sentence representations (2304 or 512 dimensions) and emotional word vectors (300 dimensions), according to the vocabulary of the emoji and hashtag dataset. These datasets respectively have 262,975 and 48,929 words in their vocabularies.

\begin{figure*}[h]
\centering
\begin{subfigure}{.5\textwidth}
  \centering
  \includegraphics[width=6cm]{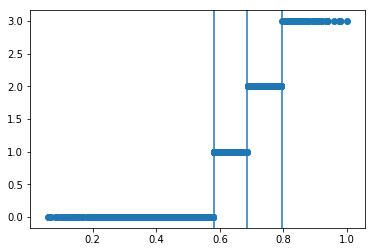}
  \caption{Fear}
  \label{fig:sub1}
\end{subfigure}%
\begin{subfigure}{.5\textwidth}
  \centering
  \includegraphics[width=6cm]{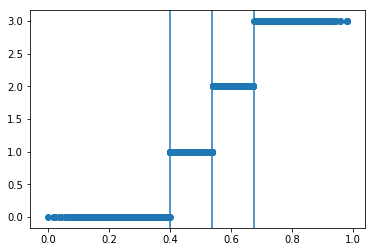}
  \caption{Joy}
  \label{fig:sub2}
\end{subfigure}
\caption{Distribution of regression labels (x-axis) and ordinal labels (y-axis) on the training dataset of Task 1a \& 2a. Class 0 for fear is distributed in [0,0.6], whereas class 0 for joy is distributed in [0, 0.4]. Vertical lines are boundaries between ordinal classes, which are used for scope mapping method}
\label{fig:reg_oc_dist}
\end{figure*}

\subsection{Regression and Ordinal Classification}
\label{sec:reg-oc}
Due to the fact that the datasets of regression tasks (EI-reg \& V-reg) and ordinal classification tasks (EI-oc \& V-oc) have the same sample sentences, we assume that regression labels are more informative than the ordinals, since they tell us the rank among the samples within the same ordinal class. Therefore, we first train a regression model and then use it to predict ordinals, rather than training a separate classifier. We later prove that this trick yields a better result in ordinal classification.

For regression, since our features are extracted from deep learning models, we find Support Vector Regression (SVR) and Kernel-Ridge Regression methods, which are effective for nonlinear features, perform better than linear methods. We tune the hyper-parameters with the given development (dev) set and later merge both train and dev set to train the final model with the best hyper-parameter found. Also, we try ensembles by averaging the final regression predictions of different methods or feature combinations to boost performance. The best groups of models are selected by the development set results of many combinations. 

Another important finding is that the mapping between the regression labels and ordinal labels are very different among emotion categories. For example in Figure \ref{fig:reg_oc_dist}, Class 0 for fear is distributed in [0,0.6], whereas class 0 for joy is distributed in [0, 0.4]. Therefore, we try to find the mapping from the regression values (continuous) to ordinal values (discrete) from the training dataset. We experiment with three different mapping:

\begin{enumerate}
\item \textit{naive mapping}: divides [0,1] into same size segments according to the number of ordinals
\item \textit{scope mapping}: finds the boundary of each segment in the training dataset (vertical lines on Figure \ref{fig:reg_oc_dist})
\item \textit{polynomial mapping}: fits a polynomial regression function from the training data and finds the closest ordinal label. 
\end{enumerate}

\subsection{Multi-label Classification}
This task of multi-label classification is different from previous tasks in that the model needs to predict the binary label for each of the 11 classes given a tweet. The task is difficult in terms of three aspects. Firstly, some of the classes have opposite emotions (such as optimism and pessimism) but may have been labeled both as true. Secondly, it is not trivial to distinguish similar emotions such as joy, love, and optimism, which will include a lot of noise in the labels and make it hard to perform classification during training. Lastly, most of the tweets are labeled with no more than 3 categories out of 11 classes, which make the labels very sparse and imbalanced (Table \ref{table:label_dist} ).

We propose to train two models to tackle this problem: regularized linear regression and logistic regression classifier chain \cite{read2009classifier}. Both models aim to exploit labels' correlation to perform multi-label classification. 

\subsubsection{Regularized linear regression model}

We formulate the multi-label classification problem as a linear regression with label distance as the regularization term. We denote the features for i-th tweet as $x_i \in R^N $ where N is the number of features and the number of categories as C. Our prediction is $y^{'}_i = W * x_i$ where $W \in R^{M*C}$ is the weight of the linear regression model. We take the following formula as loss function to minimize. The loss consists of two parts. First part aims to minimize the mean square loss between our prediction $y^{'}_i$ and ground truth label $y_i$. The second part is the regularization term to capture relationship among different emotion labels. To model the correlations among emotions, we implicitly treat each emotion category as a vertice in an undirected graph g and use Laplacian matrix of g for regularization \cite{grone1990laplacian, shahid2016pca} . 

\begin{align*}
loss &= \frac{1}{M}\sum_i^M (y' - y)^2 + \lambda y^{'T}_i L y^{'}_i \\
L &= D - A 
\end{align*}

where M is the number of samples,  $L \in R^{C*C}$ is the Laplacian matrix, $A \in R^{C*C}$ is the Euclidean matrix, $D \in R^{C*C}$ is the Degree matrix. To derive L, we first compute the co-occurrence matrix $O \in R^{C*C}$ among the emotion labels and take each row/column $O_i \in R^C$ as the representation of each emotion. Then we compute the distance matrix A  by taking the Euclidean distance of different labels. That is  $A_{ij} = (O_i - O_j)^2 $. Here, A can be regarded as the adjacency matrix of the graph g. Afterwards, we calculate the degree matrix D by summing up each row/column and making it a diagonal matrix.

\begin{table}[t]
\footnotesize
\begin{center}
\begin{tabular}{cccc}
\hline \bf{Subtasks} & \bf{1a\&2a} & \bf{3a\&4a} & \bf{5a} \\
\hline Train & 7,102 & 1,181 & 6,838 \\
Dev & 1,464 & 449 & 886 \\
Test & 4,068 & 937 & 3,259 \\
\hline
\end{tabular}
\end{center}
\caption{\label{table:dataset} Statistics of the competition dataset for all 5 subtasks}
\end{table}

\begin{table}[h]
\footnotesize
\begin{center}
\begin{tabular}{cccccccc}
\hline \bf{\# of labels} & 0 & 1 &2 &3 &4 &5 &6 \\
\hline \bf{\%} &2.9 &14.3 &40.6 &30.9 &9.6 &1.4 &0.2 \\
\hline
\end{tabular}
\end{center}
\caption{\label{table:label_dist} Number of multi-labels. Most samples have from 1-3 labels, but can have no labels or up to 6 labels. (subtask 5a)}
\end{table}

\subsubsection{Logistic regression classifier chain}
Classifier chain is another method to capture the correlation of emotion labels. It treats the multi-label problem as a sequence of binary classification problem while taking the prediction of the previous classifier as extra input. For example, when training the i-th emotion category, we take both the features of input tweet and also the 1st, 2nd, $\cdots$, (i-1)-th prediction as the input of our logistic regression classifier to predict the i-th emotion label of input tweet. We further ensemble 10 logistic regression chains by shuffling the sequence of 11 emotion labels to achieve better generalization ability.

\begin{table*}[h]
\footnotesize
\begin{center}
\begin{tabular}{|l|l|cccc|c|}
\hline \multirow{3}{*}{\bf{Features}} & \multirow{3}{*}{\bf{Regression Method}} & \multicolumn{5}{c|}{Pearson correlation (all instances)}  \\  \cline{3-7}
& & \multicolumn{4}{c|}{\bf{1a (EI-reg)}} & \bf{3a(V-reg)} \\ \cline{3-7}
 &  & Anger & Fear & Joy & Sadness & Valence  \\ 
\hline Emoji Cluster & SVR & .733 & .632 & .679 & .693 & .811 \\
Emoji Cluster & KernelRidge & .735 & .638 & .675 & .692 & .809 \\
DeepMoji & SVR & .772 & .675 & .736 & .664 & .798 \\
DeepMoji & KernelRidge & .778 & .672 & .737 & .698 & .798 \\
Emoji Cluster + EVEC & SVR & .739 & .678 & .701 & .706 & .815 \\
Emoji Cluster + EVEC & KernelRidge & .741 & .694 & .709 & .709 & .822 \\
DeepMoji + EVEC & SVR & .781 & .694 & \underline{.749} & .708 & .810 \\
DeepMoji + EVEC & KernelRidge & .779 & .702 & \underline{.754} & .710 & .813 \\
DeepMoji + feat. & SVR & \underline{.785} & .680 & .739 & .714 & .824 \\
DeepMoji + feat. & KernelRidge & \underline{.781} & .670 & .691 & .711 & \underline{.829} \\
Emoji Cluster + EVEC + feat. & SVR & \underline{.757} & \underline{.684} & \underline{.720} & \underline{.725} & \bf{\underline{.844}} \\
Emoji Cluster + EVEC + feat. & KernelRidge & \underline{.757} & \underline{.698} & \underline{.693} & \underline{.721} & \underline{.840} \\
DeepMoji + EVEC + feat. & SVR & \bf{.792} & .709 & \bf{.763} & .732 & .837 \\
DeepMoji + EVEC + feat. & KernelRidge & .790 & \underline{\bf{.716}} & .734 & \underline{\bf{.739}} & .826 \\
\hline \multicolumn{2}{|l|}{\bf{Best Ensemble}} & \underline{\bf{.811(2)}} & \underline{\bf{.728(7)}} & \underline{\bf{.773(2)}} & \underline{\bf{.753(5)}} & \underline{\bf{.860(3)}} \\
\hline
\end{tabular}
\end{center}
\caption{\label{table:reg-result} Test set results on Subtask 1a \& 3a. For 1a, separate regression models were trained for each emotion category. The number next to the best result(bold \& underlined) indicates our ranking of the competition. Underlined ones show the models that were selected for ensemble according to the dev set.}
\end{table*}

\section{Experiments \& Results}

Most of our system experiments were implemented by using PyTorch \cite{paszke2017automatic} and Scikit-learn \cite{pedregosa2011scikit}.

\subsection{Competition dataset}

SemEval-2018 Affect in Tweets (AIT) is created by human annotators through crowd-sourcing methods \cite{LREC18-TweetEmo}. Total three datasets are given: emotion intensity (with four emotion categories; Subtask 1a \& 2a), sentiment intensity (subtask 3a \& 4a), and multi-label emotion classification (subtask 5a). 

For emotion and sentiment intensity datasets, each tweet sample has both an \textit{ordinal label} (coarse; \{0,1,2,3\} for emotion, \{-3,-2,-1,0,1,2,3\} for sentiment) and real-value \textit{regression label} (fine-grained; [0,1]). For multi-label emotion classification dataset, each can have none or up to six number of multi-labels (Table \ref{table:label_dist}). 

We used the given development set to tune the hyper-parameters and select models. For the final submission, we merged the train \& development set together to retrain the model with the best hyper-parameter found (Table \ref{table:dataset}).

\subsection{Regression: Subtask 1a \& 3a}
\label{sec:reg-result}

Table \ref{table:reg-result} shows the test set results on regression tasks, Subtask 1a\&3a. We experimented with different features that we introduced before to analyze the effectiveness of each representation. For emoji sentence representations, \textit{emoji cluster} worked better on sadness and sentiment, whereas \textit{DeepMoji} outperformed in anger, fear, and joy. We presumed such difference was due to the different emoji types of the two datasets used to train each model. \textit{Emoji cluster} only used 11 classes of emojis that were clustered together, but \textit{DeepMoji} used 64 emoji classes. It may be possible clustering of emoji classes made it easy for regression models to predict the intensities in certain emotion categories, whereas some emotion categories needed more detailed representations. 

The emotional word vectors overall did help enhance the performance of the regression model for all emotion categories. This shows that emotional word vectors can serve as additional word-level information which are helpful for solving this task.

Tweet-specific features boosted the performance, notably for sentiment, since features like capital letters, emojis, elongated words, and the number of exclamation marks, could help to figure out the subtle difference of the emotion intensities.

One thing to note is that our system's rank in the fear category (7th) is relatively lower than other emotion categories. We found out from the previous literature \cite{wood2016ruder} that fear emojis were the most ambiguous, having the least correlation with human-annotated emotion labels among the six emotion categories. On the other hand, joy emojis were the most highly correlated. This may explain our best performance in the joy category and worst performance in the fear category. Future systems using emojis as a dataset may need to take this shortcoming into account.

\subsection{Ordinal Classification: Subtask 2a\& 4a}

\begin{table}[h]
\footnotesize
\begin{center}
\begin{tabular}{|l|r|ccc|}
\hline \multicolumn{2}{|c|}{} & \multicolumn{3}{c|}{Pearson (all instances)} \\ \cline{3-5}
\multicolumn{2}{|c|}{\bf{Task}} & Naive & Scope & Poly \\
\hline \multirow{5}{*}{2a (EI-oc)} & Anger & .654 & .664 & \bf{.704(2)} \\
 & Fear & .498 & .562 & \bf{.570(*)} \\
 & Joy & .632 & \bf{.720(1)} & .712 \\
 & Sadness & .645 & \bf{.697(*)} & .692 \\
\hline 4a (V-oc) & Valence & .813 & .816 & \bf{.833(2)} \\
\hline
\end{tabular}
\end{center}
\caption{\label{table:oc-result} Test set results on Subtask 2a \& 4a. The predictions of the best regression models are mapped into ordinal predictions. The number next to the best result(bold \& underlined) indicates our ranking of the competition. (*) indicates better results that we acquired after our final submission}
\end{table}

As mentioned in Sec \ref{sec:reg-oc}, we used our best regression model to also predict ordinal labels. Since each emotion category has a different distribution of regression labels and ordinal labels, we experimented three different mappings, \textit{naive mapping}, \textit{scope mapping}, and \textit{polynomial mapping}. Using the training set, we found the ideal mapping function to match the regression predictions and the ordinal predictions. 

\begin{figure}[t]
\centering
\includegraphics[width=7cm]{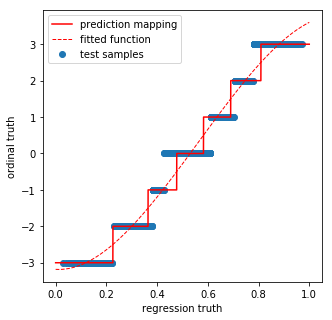}
\caption{Plot of test labels and the mapping function derived from the training set. A polynomial  function is fitted to map the regression predictions into ordinal predictions}
\label{fig:v_test_map}
\end{figure}

Test set results (Table \ref{table:oc-result}) on ordinal classification show that our mapping methods are indeed much more effective. For anger, fear, and sentiment categories, polynomial mapping performed the best, whereas scope mapping outperformed for joy and sadness categories. With our method, we achieved higher ranks in ordinal classification tasks (2a \& 4a), placed both in 2nd. Figure \ref{fig:v_test_map} shows how a cubic function is fitted to find the mapping between regression labels and ordinal labels. 

Additionally, we report some results better than the final submission. The change is due to a new model selection strategy. For the final submission, we searched for the optimal pair of regression model \& mapping method by looking at the ordinal classification results on the development set. However, it turned out that always using the best ensemble prediction and then searching for the optimal mapping method with respect to the development set was better.

\subsection{Multi-label Classification: Subtask 5a}
We found the best hyper-parameters by evaluating on our development set. We initialized the weight matrix W with a normal distribution of standard deviation of 0.1.  We used gradient descent to optimize this function and set the learning rate to 1.0. The optimal $\lambda$ we found was -0.0001. 

We found that regularized linear regression model was always better than classifier chain model. The ensemble of classifier chain and regularized linear regression of both features combination(underlined elements in Table  \ref{table:clf-result}) achieved best performance than any single model (Table \ref{table:clf-result}).

\begin{table}[t]
\footnotesize
\begin{center}
\begin{tabular}{|l|cc|}
\hline \bf{Features} & \bf{CC} & \bf{RLR}  \\
\hline Emoji Cluster & .528 & .545  \\
DeepMoji & .532 & .552 \\
Emoji Cluster + EVEC & .545 & .558  \\
DeepMoji + EVEC & .544 & .558   \\
Emoji Cluster + EVEC + f & \underline{.546}  & \underline{.558} \\
DeepMoji + EVEC + f & \underline{.550}  & \underline{.562}  \\
\hline
\bf{Best ensemble} & \multicolumn{2}{|c|}{\underline{\bf{.576(3)}}} \\ 
\hline
\end{tabular}
\end{center}
\caption{\label{table:clf-result} Test set results on Subtask 5a. The competition metric is Jaccard index.}
\end{table}

\begin{table}[t]
\footnotesize
\begin{center}
\begin{tabular}{|c|cc|cc|}
\hline & \multicolumn{2}{|c|}{\bf{Gender}} & \multicolumn{2}{|c|}{\bf{Race}} \\
\hline & \underline{\bf{Ours}} & Avg &   \underline{\bf{Ours}} & Avg \\
\hline Anger &  \underline{\bf{0.5\%}} & 0.1\% &  \underline{\bf{1\%}} & 0.4\%  \\
Fear & \underline{\bf{-0.9\%}} & -0.3\% & \underline{\bf{3.3\%}} & 0.5\% \\
Joy & \underline{\bf{-1.2\%}} & 0.4\% & \underline{\bf{-0.9\%}} & -0.7\% \\
Sadness & \underline{\bf{0\%}} & 0.2\% & \underline{\bf{1.3\%}} & 0.8\% \\
Valence & \underline{\bf{-0.6\%}} & 0.5\% & \underline{\bf{-1\%}} & -0.6\% \\
\hline
\end{tabular}
\end{center}
\caption{\label{table:bias-result} Average differences of the system's bias. Gender difference is from female to male, and race differences is from African American names to European American names (sign of the percentage indicates the direction). ``Ours`` indicate the bias of our system, and ``Avg`` is the average of the biases of all systems from the competition.}
\end{table}

\subsection{Analysis of system's gender/racial biases}

In this year's competition, the organizers gave out a mystery test set that was included in the regression tasks (subtask 1a \& 3a). At the end of the evaluation period, they announced that these were set of pair sentences that differ only in the subject's or object's gender or racial names (See the task paper \citet{SemEval2018Task1} for details). It turned out that our system also included some biases like most other systems did, but fairly small, less than 1.5\% for gender bias and 3.5\% for racial bias (Table \ref{table:bias-result}). We believe that this is an interesting experiment and look forward to discussing more about the issue during the workshop.

\begin{table}[h]
\footnotesize
\begin{center}
\begin{tabular}{|c|c|c|}
\hline \bf{Subtask} & \bf{System} & \bf{Score(rank)} \\
\hline
\multirow{4}{*}{\bf{1a EI-reg}}  & SeerNet & .799(1) \\
& NTUA-SLP & .776(2) \\
& \underline{\bf{PlusEmo2Vec}} & .766(3) \\
& psyML & .765(4) \\
\hline
\multirow{3}{*}{\bf{2a} EI-oc}  & SeerNet & .695(1) \\
& \underline{\bf{PlusEmo2Vec}} & .659(2) \\
& psyML & .653(3) \\
\hline
\multirow{4}{*}{\bf{3a V-reg}}  & SeerNet & .873(1) \\
& TCS Research & .861(2) \\
& \underline{\bf{PlusEmo2Vec}} & .860(3) \\
& NTUA-SLP & .851(4) \\
\hline
\multirow{3}{*}{\bf{4a} V-oc}  & SeerNet & .836(1) \\
& \underline{\bf{PlusEmo2Vec}} & .833(2) \\
& Amobee & .813(3) \\
\hline
\multirow{4}{*}{\bf{5a E-c}}  & NTUA-SLP & .588(1) \\
& TCS Research & .582(2) \\
& \underline{\bf{PlusEmo2Vec}} & .576(3) \\
& psyML & .574(4) \\
\hline

\end{tabular}
\end{center}
\caption{\label{table:ranking} Official final scoreboard on all 5 subtasks that we participated. Scores for Subtask 1-4 are macro-average of the Pearson scores of 4 emotion categories and 5 is Jaccard index. About 35 participants are in each task.}
\end{table}

\section{Conclusion}

In this paper, we explored a couple of different methods to find good representations of emotions inside tweets for solving 5 subtasks of predicting emotion/sentiment intensity and emotion labels. We used external datasets, which were much larger than the competition dataset but distantly labeled with emojis and \#hashtags, to exploit the transferred knowledge to build a more robust machine learning system to solve the task. We avoided using traditional NLP features like linguistic features and emotion/sentiment lexicons by substituting them with continuous vector representations learned from huge corpora. 

We performed experiments to show that emoji sentence representations and emotional word vectors trained from neural networks can be used with tweet-specific features as input for other traditional regression models, such as SVR and Kernel Regression, to solve the task of regression and ordinal classification. We proved the effectiveness of finding the mapping of the relationship between regression and ordinal labels from the training set to perform ordinal classification. Moreover, we tried using classifier chain and regularized logistic regression to deal with multi-label classification. 

As a final official result (Table \ref{table:ranking}), our system ranked among the top three in every subtask of the competition we participated. For future work, we want to work further on employing these emotion representations on other tasks, such as text generation, while we gather more data and improve the model to train the representations.
\newpage

\section*{Acknowledgments}
This work is partially funded by ITS/319/16FP of Innovation Technology Commission, HKUST 16214415 \& 16248016 of Hong Kong Research Grants Council, and RDC 1718050-0 of EMOS.AI.
 \\

\bibliographystyle{acl_natbib}
\bibliography{semeval2018}

\end{document}